\pdfoutput=1

\documentclass[11pt]{article}

\usepackage{ACL2023}

\usepackage{times}
\usepackage{latexsym}

\usepackage[T1]{fontenc}

\usepackage[utf8]{inputenc}

\usepackage{microtype}

\usepackage{inconsolata}

\usepackage{algorithm}
\usepackage{algorithmic}
\usepackage{xcolor}
\usepackage{color}
\usepackage{makecell}
\usepackage{amsmath}
\usepackage{booktabs}
\usepackage{multirow}
\usepackage{multicol}
\usepackage{subfigure}

\definecolor{nmgray}{RGB}{229,229,229}
\definecolor{underlinegray}{RGB}{197,197,197}

\definecolor{introblue}{RGB}{0,176,240}
\definecolor{introgreen}{RGB}{0,203,134}
\definecolor{introgreen2}{RGB}{139,243,206}
\usepackage{tcolorbox}

\usepackage{adjustbox}

\newtcolorbox{mybox}[2][]{
width=\columnwidth,
colback = nmgray!75!white, 
colframe = nmgray!75!white, 
boxsep=0pt,left=10pt,right=10pt,top=0pt,bottom=0pt,
fontupper=\linespread{0.9}\selectfont,
title=#2,#1}

\usepackage{newfloat}
\usepackage{listings}
\DeclareCaptionStyle{ruled}{labelfont=normalfont,labelsep=colon,strut=off} 
\newcommand{\vpara}[1]{\vspace{0.05in}\noindent \textbf{#1 }}

\title{Building Emotional Support Chatbots in the Era of LLMs}

\author{Zhonghua Zheng$^{1}$ \quad 
Lizi Liao$^{2}$ \quad
Yang Deng$^{3}$ \quad
Liqiang Nie$^{1}$ \\[1ex]
  $^{1}$ Harbin Institute of Technology, Shenzhen \quad $^{2}$ Singapore Management University \\
  $^{3}$ National University of Singapore \\[1ex]
{\tt \small \{polang1999, liaolizi.llz\}@gmail.com, ydeng@nus.edu.sg, nieliqiang@gmail.com}}

\begin{document}
\maketitle

\begin{abstract}
While emotional support in conversational scenarios offers societal benefits, limited data and non-standardized training impede its application. This work endeavors to navigate these challenges by harnessing the capabilities of Large Language Models (LLMs). We introduce an innovative methodology that synthesizes human insights with the computational prowess of LLMs to curate an extensive emotional support dialogue dataset. Our approach is initiated with a meticulously designed set of dialogues spanning diverse scenarios as generative seeds. By utilizing the in-context learning potential of ChatGPT, we recursively generate an \textbf{ExT}ensible \textbf{E}motional \textbf{S}upport dialogue dataset, named \textbf{ExTES}. Following this, we deploy advanced tuning techniques on the LLaMA model, examining the impact of diverse training strategies, ultimately yielding an LLM meticulously optimized for emotional support interactions. An exhaustive assessment of the resultant model showcases its proficiency in offering emotional support, marking a pivotal step in the realm of emotional support bots and paving the way for subsequent research and implementations. The dataset and codes are available here\footnote{https://anonymous.4open.science/r/ExtESC-2761/}.
\end{abstract}

\begin{figure*}[ht]
    \centering
    \includegraphics[width=0.90\linewidth]{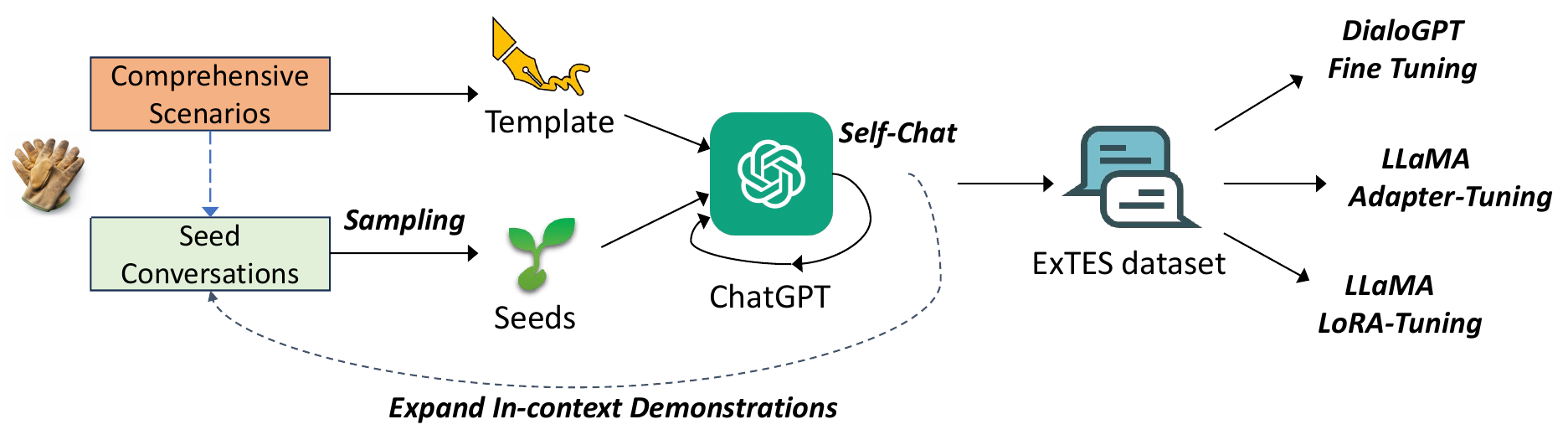}
    \caption{The pipeline for collecting the ExTES conversation dataset via our proposed extendable looping scheme. Based on the dataset, we benchmark and analyze the effect of fine-tuning the chat model with various techniques.}
    \label{Figure1}
\end{figure*}

\section{Introduction}
In today's interconnected world, the integration of emotional support into various conversational contexts holds profound significance \cite{Li_2021}. Emotional support conversations play a pivotal role in fostering empathy, understanding, and overall well-being among individuals. These conversations create safe spaces where emotions can be openly expressed and validated, allowing individuals to share their joys, sorrows, fears, and challenges. By forging deep connections and nurturing a sense of community, emotional support bots contribute to a more caring and supportive society.

Despite the undeniable importance of Emotional Support (ES) dialogue systems, the real-world applications are largely hindered by the glaring lack of large-scale well-annotated datasets \cite{sun2021psyqa}. 
Most of existing studies in emotional support conversations prioritize dataset collection from online sources, such as stress-related Twitter interactions \cite{Medeiros}, mental health reddits \cite{sharma-etal-2020-computational}, and online support groups \cite{Hosseini}.
However, most of these conversations are asynchronous and limited to single-turn interaction scenarios. Contrarily, \citet{huang2021} introduced the ESConv dataset via questionnaires, highlighting quality collection and multi-turn conversation. Yet, its constraints stem from its modest size and lack of extensive strategy annotations and scenario variety, likely due to the substantial costs associated with its compilation. 

Moreover, this domain lacks standardized training methodologies. 
Earlier works typically settled for small, manually annotated datasets, which were deeply anchored in traditional approaches like face-to-face therapy \cite{gibson2016deep} and motivational interviewing \cite{perez2017understanding}. The spotlight then was primarily on empathy detection and discerning predictive rationales \cite{rashkin-etal-2019-towards,sharma2020computational}.
A shift emerged with innovative prototypes like  the Emotional chatting machine \cite{zhou2018emotional}, focusing on emotion-centric dialogue generation. Simultaneously, there's a surge in studies aimed at controlled text generation. \citet{hu2017toward} presented a sentiment and tense-influenced model, while \citet{ghosh2017affect} and \citet{cagan2017data} respectively harnessed affective contexts and grammar-focused techniques. Recent evaluations by \citet{huang2021} on platforms like BlenderBot and DialoGPT and GPT-driven explorations by \citet{chowdhury2022novelty} mark the field's evolution. 
Nonetheless, ranging from small-scale models such as RNNs to medium-sized ones like GPT-2, a dominant leaning towards sequence-to-sequence models persists. 

Recently, Large Language Models (LLMs) have emerged as an epicenter of innovation, exhibiting remarkable generative prowess \cite{pmlr-v202-aher23a}. Notably, ChatGPT registers commendable feats across a gamut of NLP challenges and benchmarks \cite{tan2023chatgpt}. A salient feature introduced by this model is ``In-Context Learning''. This paradigm allows few-shot samples as prompts to facilitate model learning and nuanced generation \cite{ji2023exploring}. Parallel to this, initiatives like Stanford Alpaca \cite{alpaca} and Vicuna \cite{vicuna2023} have sought to emulate ChatGPT's success, employing data from GPT-3.5 and subsequently refining LLaMA \cite{2023llama}. Collectively, these advancements illuminate the tantalizing potential: deploying LLMs for crafting emotional support dialogues and optimizing the LLaMA architecture could conceivably address the twin challenges of data scarcity and training methodology refinement.

Correspondingly, we present a novel study that focuses on developing emotional support dialogue systems with the help of LLMs. 
As illustrated in Figure \ref{Figure1}, the proposed architecture address the pressing concerns of data scarcity  by capitalizing on LLMs while sculpting a nuanced emotional support-centric LLM.
Initially, we synergize human expertise with LLMs' capabilities, forging the creation of an \textbf{ExT}ensible \textbf{E}motional \textbf{S}upport dialogue dataset, termed as \textbf{ExTES}. Our approach encompasses the craft of a meticulous and encompassing set of emotion support dialogues, spanning various scenarios and enriched with strategic annotations. These dialogues serve as primordial seeds, prompting ChatGPT to recursively generate expansive dialogues, tapping into its adept in-context learning capabilities. The subsequent phases delve into the exploration of diverse fine-tuning methodologies aimed at refining our emotional support chatbot model. We undertake exhaustive 
evaluations via automatic and human evaluations, assessing its proficiency in rendering emotional support. The analyses affirm that (1) The ExTES, when sculpted in concert with LLMs, notably transcends ESConv's quality across multiple dimensions, laying a fertile groundwork for subsequent inquiries and deployments. (2) The judicious application of parameter-efficient tuning techniques on LLaMA \cite{2023llama} emerges as an optimal blueprint for the evolution of proficient emotional support chatbots.

To summarize, our main contributions in this paper are as follows:
\begin{itemize}
\item We innovatively leverage the generative capabilities of LLMs to generate an extensible emotional support dialogues dataset, ExTES, 
with comprehensive scenarios and strategies, which is released for building robust and generalizable emotional support systems.
\item We investigate different fine-tuning strategies to endow LLaMA with effective and flexible emotional support capabilities. 
\item 
Extensive evaluations validate the effectiveness and significant values of the ExTES dataset.
The successful integration of emotional support dialogue with LLMs can positively impact mental health counseling, social interactions, customer service, and various other domains, contributing to a more compassionate and supportive society. 
\end{itemize}

\section{Related Work}
\paragraph{Emotional \& Empathetic Conversation.} 
Emotion's role in building engaging dialogue systems has been extensively studied \cite{zhou2018emotional,li-etal-2017-dailydialog,zhou-wang-2018-mojitalk,Huber2018,huang2020challenges}. \cite{zhou2018emotional} proposed the Emotional Chatting Machine (ECM) to generate emotional responses based on pre-specified emotions, accurately expressing designated emotions. However, it's essential to differentiate emotional chatting from emotional support (ES). While emotional chatting expresses emotions like happiness or sadness, ES aims to reduce user emotional distress through proactively guiding the conversation \cite{liao2023proactive,deng2023prompting} and leveraging proper support skills. 
Empathetic responding is another related task in emotional conversation research \cite{rashkinetal2019towards,linetal2019moel,majumderetal2020mime,zandie2020emptransfo,sharma2020computational,zhongetal2020towards,sun2021psyqa}. It focuses on understanding users' feelings \cite{liao2021dialogue} and generating appropriate replies. 
\citet{rashkinetal2019towards} argued that recognizing interlocutors' emotions enables more empathetic responses. Empathetic responding is a critical aspect of effective ES but is just one component within the broader framework.
Besides empathetic responding, emotional support conversations further require the capability of dialogue strategy learning~\cite{emnlp22-esc,acl22-esc,ijcai22-esc,acl23-esc} to address users' problems and provide assistance in coping with difficulty.
In other words, the key challenge in emotional support conversations is to make strategic decision for handling various emotional issues.

\begin{table*}[!htp]
\centering
\normalsize
\renewcommand*{\arraystretch}{1.2}
\resizebox{\textwidth}{!}
{ 
\begin{tabular}{lcc|lcc}
\toprule
\textbf{Category} & \textbf{Dialogues} & \textbf{Proportion} & \textbf{Category} & \textbf{Dialogues} & \textbf{Proportion}\\
\midrule
Breakups or Divorce& 710 &6.3\%   &Navigating Gender Identity and Transitioning &202 &1.8\%\\

Conflicts or Communication Problems&1,109 &9.9\%   &Moving to a New City or Country &202 &1.8\%\\

Communication Challenges& 1,008 &9.0\%  &Career Transitions &202 &1.8\%\\

Coping with the Death of a Loved One&593 &5.3\%  &Parenthood and Parenting Challenges &202 &1.8\%\\

Dealing with the Loss of a Pet&601 &5.4\%  &Low Self-Esteem or Lack of Confidence &302 &2.7\%\\

Work-related Stress and Burnout&403 &3.6\%  &Body Image Concerns and Eating Disorders &101 &0.9\%\\

Financial Worries and Uncertainty&403 &3.6\%  &LGBTQ+ Identity &101 &0.9\%\\

Unemployment-related Stress&403 &3.6\%  &Cultural Identity and Belonging &101 &0.9\%\\

Academic Stress&403 &3.6\%  &Academic Stress or Pressure &202 &1.8\%\\

Spirituality and Faith&202 &1.8\%  &Job Loss or Career Setbacks &202 &1.8\%\\

Managing Bipolar Disorder&202 &1.8\%  &Parenting Challenges and Parental Guilt &202 &1.8\%\\

Anxiety and Panic&202 &1.8\%  &Sibling Rivalry or Family Conflict &403 &3.6\%\\

Depression and Low Mood&403 &3.6\%  &Surviving and Recovering from Physical or Emotional Abuse &101 &0.9\%\\

Adjusting to a New Job or Role&302 &2.7\%  &Healing from Sexual Assault or Domestic Violence &101 &0.9\%\\

Chronic Illness or Pain Management&302 &2.7\%  &Post-Traumatic Stress Disorder (PTSD) &101 &0.9\%\\

Coping with a Diagnosis or Medical Treatment&202 &1.8\%  &Healing from Abuse &202 &1.8\%\\

Caregiver Support&202 &1.8\%  &Addiction and Recovery &202 &1.8\%\\

Finding Meaning and Purpose in Life&202 &1.8\%  &Support for Loved Ones or Friends &202 &1.8\%\\
\bottomrule
\end{tabular}
}
\caption{Statistics of all 36 emotional support scenarios covered in our ExTES dataset.\\}
\label{scenarios}
\end{table*}

\paragraph{Language Models for Conversations.}
The field of language models for conversations has seen remarkable progress recently \cite{liao2023proactive,ye2022structured}, with various models demonstrating impressive capabilities in open-domain dialogue, such as DialoGPT \cite{zhang2020dialogpt}, Meena \cite{2020arXiv200109977A}, LaMDA \cite{2022arXiv220108239T}, etc.
ChatGPT \cite{instructgpt} optimized language models for chat using Reinforcement Learning with Human Feedback, resulting in human-like chat abilities, which was further enhanced into GPT4 with advanced reasoning and multi-modal capabilities in conversations \cite{liao2021mmconv,ye2022reflecting}. 
Meanwhile, researchers aimed to replicate ChatGPT's success with open-source foundation models. For example, Stanford Alpaca \cite{alpaca} used Self-Instruct \cite{wang2023selfinstruct} to collect GPT-3.5 data and fine-tuned the LLaMA model \cite{2023llama}. Vicuna \cite{vicuna2023} fine-tuned LLaMA on a dialogue corpus from sharegpt.com, enhancing its conversational capabilities. Amidst this evolving landscape, our work contributes by exploring techniques to improve the quality, contextuality, and empathy of emotional support conversations.

\paragraph{Related Datasets for Emotional Support.} 
Early studies on emotional support conversations typically focused on collecting and annotating datasets derived from social media \cite{Medeiros}, online forums \cite{sharmaetal2020computational,Hosseini}, or psychotherapy video transcripts \cite{sigdial20-counselling-data}. However, the emotional support strategy is overlooked and the dialogue quality is unsatisfactory in these datasets. To this end, \citet{huang2021} constructed an Emotional Support Conversation dataset, ESConv, which carefully designs the process of data collection and devises multiple mechanisms to ensure the effectiveness of emotional support strategies in conversations. However, the types of emotional support strategies and the categories of dialogue backgrounds are limited, which is difficult to be generalized to complex situations in reality.

\section{Dataset Collection}
In this section, we elucidate our approach to constructing a sophisticated multi-turn emotional support chat corpus harnessing ChatGPT (gpt-3.5-turbo). The process bifurcates into two stages. Initially, we delineate comprehensive emotional support scenarios, encompassing response strategies, and meticulously curate exemplar dialogues from extant datasets and online platforms, emphasizing richness and relevance. Subsequently, we leverage ChatGPT to generate an extended set of dialogues using these seed exemplars, followed by a manual correction process. The refined dialogues will then be reintroduced into the model as supplementary exemplars to further enrich the dataset in an iterative fashion. Notably, our strategy offers a marked reduction in human labor, harmonizing efficiency with dataset integrity.

\subsection{Comprehensive Scope and Strategies}
To ensure the diversity and broad coverage of emotional support conversations, it is important to include comprehensive emotional support conversation scenarios and response strategies. Drawing upon a wealth of literature on psychological counseling \cite{2003syrategies} and insights from previous emotional support research \cite{reblin2008social,zmab005, SHENSA202038, graham2019artificial}, we create a comprehensive set of emotionally diverse scenarios. Additionally, we referred to emotional support scenes from related works, expanding and refining our collection. In the end, we obtain 36 emotionally impactful scenarios (listed in Table \ref{scenarios}), detailed in Appendix D. 
These scenarios encompass a wide array of everyday life situations, catering to the varied emotional needs of users. Unlike the limited five scenarios in ESConv \cite{huang2021}, our expanded range now provides a richer array of contexts for emotional support interactions. This expansion has been instrumental in refining the content and scope for providing emotional support.

Similarly, inspired by \cite{Hill1999HelpingSF,world2020mental}, we compile 16 emotional support strategies (Table \ref{strategies}). Compared with the eight strategies in ESConv, our emotional support strategies are richer and more conducive to providing users with targeted suggestions.

\subsection{In-context Examples and Collection}
We initiated our data collection process by manually constructing 87 seed dialogues. These seed dialogues were sourced from real emotion support datasets, including ESConv\cite{huang2021}, ETMHS\cite{sharma2020empathy}, and Reddit\cite{Yeh2021ADD} datasets. Additionally, we used web crawling to supplement our data collection efforts for certain missing scenarios. By incorporating real cases from these datasets and web-crawled materials, we ensured a diverse and representative set of emotional support dialogues.
Each of the 36 scenarios we identified has at least two seed dialogues, ensuring sufficient coverage. The quality of dialogues can be guaranteed after careful manual correction on the dialogues and response strategy labeling.

\subsection{Human Review and Iteration}
With meticulously curated exemplar dialogues, we prompt ChatGPT to generate expansive dialogues via self-chat, which has been shown effective in various dataset solicitation works \cite{xu2023baize}. It involves utilizing ChatGPT to generate answers for the user and the emotional counseling assistant in the dialog format we set. Specifically, we design a template (shown in Appendix C
) to define the format and requirements, allowing the API to continuously generate transcripts of both sides of the conversation, and annotate appropriate emotional support strategies. The dialogue takes the seed dialogue as an example and outputs complete dialogues according to given scenarios. Specifically, We first filled the 87 seed dialogues into the template, collected a total of 1k dialogues with different dialogue scenarios through self-chat, and then continued to collect data by replacing the seed dialogues with these 1k dialogues. The whole process is iterative and expansive, \textit{i.e.} can be expanded easily with new seeds and new scenarios.

Note that the whole collection process heavily rely on ChatGPT's self-chat, hence the quality of generated dialogues is our major concern. Although the format and requirements of the output dialogues are clearly specified in the template we defined, we still observe some data format errors, duplication, and failures to meet other requirements. To ensure the quality, we conduct human review on the collected dialogues in every iteration and apply manual correction process. However, we notice that the human intervention requirement is rather limited as compared to existing conversation dataset collection methods such as questionnaire \cite{huang2021} or crowed-sourcing \cite{budzianowski2018multiwoz}. For instance, we note that fewer than 10\% of the generated conversations require human adjustments. Typical discrepancies include formatting issues, absent response strategies, strategy rectifications, and the inclusion of specific details. Conversations needing extensive revisions are directly discarded. After the screening and modification process, we finally obtain around 11k conversations that meet the requirements to form ExTES. This process costs about \$210 to call OpenAI's API\footnote{https://platform.openai.com/docs/api-reference}.

\begin{table}[!htp]
      \vspace{2mm}
\centering
\normalsize
\renewcommand*{\arraystretch}{1.2}%
\resizebox{0.48\textwidth}{!}{
\begin{tabular}{lcc}
\toprule
\textbf{Category} & \textbf{Dialogues} & \textbf{Proportion}\\
\midrule
Reflective Statements (RS)& 14,560 & 14.8\% \\
Clarification (Cla)& 2,898& 2.9\%\\
Emotional Validation (EV)& 19,367& 19.8\%\\
Empathetic Statements (ES)& 8,482& 8.7\%\\
Affirmation (Aff)& 16,539& 16.9\%\\
Offer Hope (OH)& 4,665& 4.8\%\\
Avoid Judgment And Criticism (AJC)& 1,767& 1.8\%\\
Suggest Options (SO)& 6,079& 6.2\%\\
Collaborative Planning (CP)& 3,534& 3.6\%\\
Provide Different Perspectives (PDP)& 3,322& 3.4\%\\
Reframe Negative Thoughts (RNT)& 2,050 & 2.1\%\\
Share Information (SI)& 3,181& 3.3\%\\
Normalize Experiences (NE)& 2,403& 2.6\%\\
Promote Self-Care Practices (PSP)& 2,686& 2.7\%\\
Stress Management (SM)& 2,474& 2.5\%\\
Others (Oth)& 3887& 3.9\%\\
\midrule
Overall& 97,893& 100\%\\
\bottomrule
\end{tabular}
}
\caption{Statistics of response strategies used in ExTES.\\}
\label{strategies}
\vspace{-0.2cm}
\end{table}

\begin{table}[ht]
\centering
\resizebox{0.45\textwidth}{!}{
\begin{tabular}{lccc}
\toprule
\textbf{Category} & \textbf{ExTES} & \textbf{ESConv}\\
\midrule
Dialogues& 11,177&1,053\\
Utterances& 200,393& 31,410\\
Avg. length of dialogues& 18.2& 29.8\\
Avg. length of utterances& 26.0& 17.8\\
Num. of support strategise& 16& 8\\
Num. of scenarios& 36& 5\\
\bottomrule
\end{tabular}
}
\caption{The statistics of our ExTES vs. ESConv.\\}
\label{statistics}
\end{table}

\section{Dataset Characteristics and Quality}
\vpara{General Statistics}
Our resultant dataset ExTES comprises 11,177 dialogues in total, with specifics provided in Table \ref{statistics}. On average, a dialogue contains 18.2 utterances. While users often display negative sentiments, the assistants lean towards positive expressions, offering emotional support. An illustrative dialogue is displayed in Figure 4 
in Appendix A.

The length of these dialogues, averaging 18.2 utterances, suggests that delivering effective emotional support frequently necessitates multiple conversational turns. This is considerably more than what's observed in prior datasets on emotional chatting \cite{zhou-wang-2018-mojitalk} and empathetic dialogue \cite{rashkin-etal-2019-towards}. While our dialogues are shorter on average than those in ESConv (29.8 utterances on average), they have a more substantial average utterance length (26.0 versus ESConv's 17.8). This suggests our dialogues are content-rich. Detailed statistics for other annotations can be found in Table \ref{scenarios} (highlighting ES scenarios) and Table \ref{strategies} (detailing ES strategies). Notably, the most prevalent emotional challenges arise from communication issues, a trend we identified early in our seed data collection. Following closely are problems stemming from work-related pressures or unemployment, possibly exacerbated by global economic downturns.

\begin{table}[ht]
  \centering
    \scalebox{0.8}{
    \begin{tabular}{lccc}
    \toprule
    & \textbf{Crowd-sourced} & \textbf{ExTES} & $\kappa$ \\
    \midrule
    \textbf{Informativeness} & 2.39 & \textbf{2.53} & 0.51\\
    \textbf{Understanding} & \textbf{2.64}& 2.52 & 0.46\\
    \textbf{Helpfulness} & 2.48 & \textbf{2.61} & 0.44 \\
    \textbf{Consistency} & \textbf{2.75} & 2.67 & 0.39 \\
    \textbf{Coherence} & 2.38 & \textbf{2.45} & 0.52 \\
    \bottomrule
    \end{tabular}%
    }
    \caption{
  Human evaluation of ExTES quality.
  The scores (from 0 to 3) are averaged over all the samples rated by three annotators. $\kappa$ denotes Fleiss' Kappa \cite{Fleiss1971}, indicating fair to moderate inter-annotator agreement ($0.2 < \kappa < 0.6$).
  }
  \label{tab:quality}%
\end{table}%
\vpara{Dialogue Quality Evaluation}
We further validate the quality of ExTES via conducting comprehensive human evaluation and comparing the collected ExTES part with our seed dialogues (Crowd-sourced).
Referring to \cite{huang2021}, we adopted the following metrics to evaluate the quality of our augmented dialogues.
\textbf{Informativeness} evaluates the extent to which the individual seeking assistance elaborates on their emotional distress.
\textbf{Understanding} assesses the level of comprehension the supporter has of the individual's experiences and feelings.
\textbf{Helpfulness} determines the degree to which the supporter is successful in alleviating the emotional discomfort of the individual and improving their mood.
Moreover, we also evaluate the general quality of the dialogue.
\textbf{Consistency} checks if the participants' behavior is in line with their roles, and if there is no contradiction in a participant's behavior.
\textbf{Coherence} assesses if the conversation stays focused and in-depth, and if transitions between topics are smooth.
All the metrics are rated with the four-level Likert scale \cite{allen2007likert} ranging from 0 to 3 (higher is better). We recruited 50 college students from different majors to serve as annotators for this project. We randomly selected 100 dialogue examples from ExTES and 50 examples from our seed dialogues. At least two different annotators rated each dialogue example.

As shown in Table~\ref{tab:quality}, it demonstrates that our method can generate high-quality emotional support dialogues as in ExTES. Dialogues collected by our method show similar evaluation scores compared to crowdsourced seed dialogues. It is even better than seed dialogues in terms of Informativeness and Helpfulness. According to our observation, this might be because the answers generated by ChatGPT tend to have more substantial and complete content.

\vpara{Strategy Distribution}
In our study, we also hope to comprehend the distribution of strategies adopted by ChatGPT in generating responses at various stages of the dialogue. For the purpose of this analysis, we considered a conversation with N utterances in total, where the $k$-th utterance is from the assistant and adopts the strategy $S$. The position of this utterance in the conversation is referred to as the conversation phases and is represented as $k/N$. We evenly divide the conversation progress into four phases.

\begin{figure}[ht]
    \centering
    \includegraphics[width=1\linewidth]{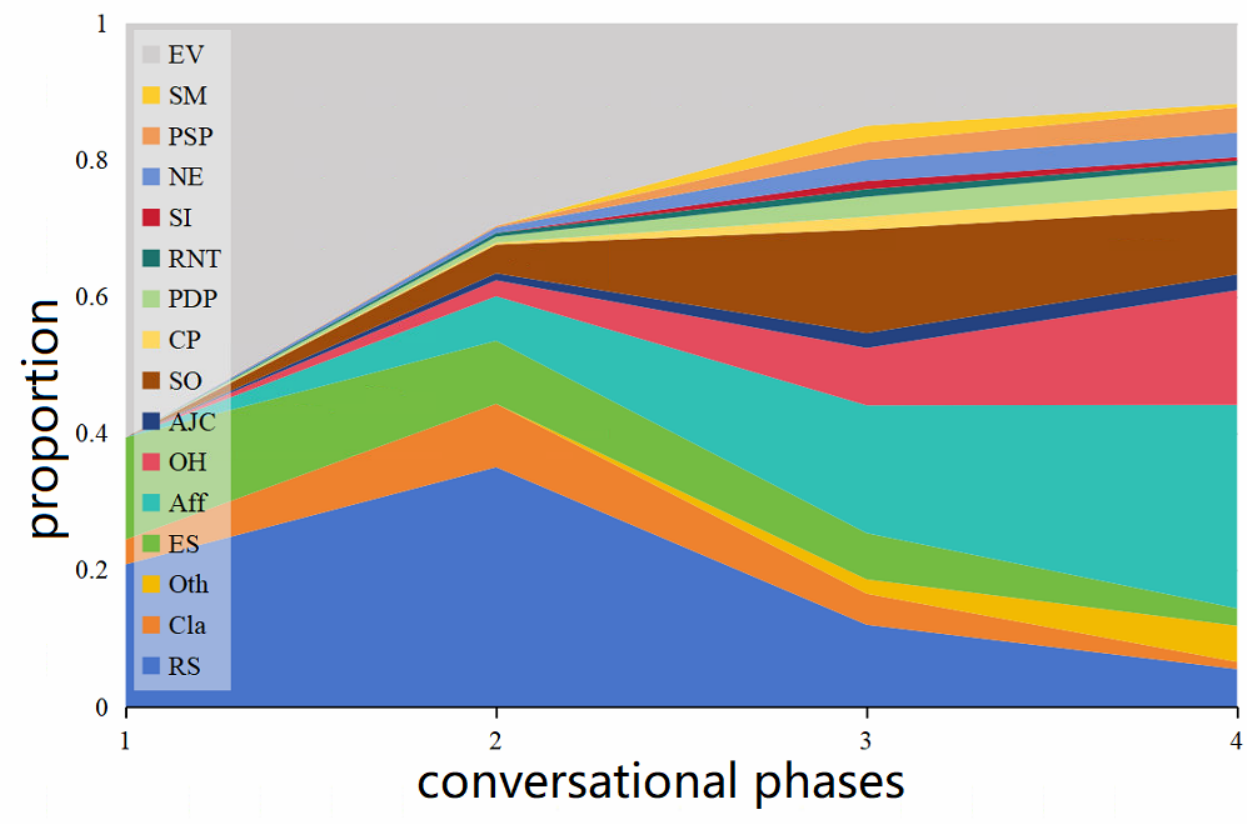}
    \caption{The distribution of strategies at different phases.}
    \label{Figure2}
\end{figure}

We examined all the dialogues in our dataset and tallied the proportions of different strategies adopted within these four intervals. The resultant data points provided a snapshot of the strategy distribution at four phases along the conversation progress. As shown in Figure \ref{Figure2}, we can observe different trend in the distribution of ES strategies across the different phases of the conversation.

\vpara{Strategy Transition}
We present the top-5 most frequent strategy transitions with 3-5 hops in Table \ref{tab:transition}. These transitions indicate that assistants usually ask questions and explore the help-seekers situations before comforting the help-seek users. Emotional support assistants usually first understand the cause of the user's distress and then say some words of comfort or express sympathy for the user's experience. This is generally as expected. It also might not be wise enough to make actionable suggestions at the beginning of the whole dialogue.

\begin{table}[t]
      \vspace{1mm}
  \centering
  \scalebox{0.82}{
    \begin{tabular}{clr}
        \toprule
              & \textbf{Strategy Transition} & \textbf{Proportion} \\
        \midrule
        \multirow{5}[0]{*}{\textbf{3-Hop}} & EV $\to$ RS $\to$ EV & 17.19 ‰ \\
              & EV $\to$ RS $\to$ SO & 16.23 ‰ \\
              & EV $\to$ RS $\to$ ES & 14.49 ‰ \\
              & RS $\to$ EV $\to$ SO & 11.03 ‰ \\
              & EV $\to$ ES $\to$ RS & 9.75 ‰ \\
        \midrule
        \multirow{5}[0]{*}{\textbf{4-Hop}} & EV $\to$ RS $\to$ ES $\to$ SO & 7.08 ‰ \\
              & EV $\to$ RS $\to$ SO $\to$ Aff & 6.61 ‰ \\
              & EV $\to$ ES $\to$ RS $\to$ NE & 6.04 ‰ \\
              & RS $\to$ Aff $\to$ ES $\to$ RS & 5.27 ‰ \\
              & EV $\to$ RS $\to$ SO $\to$ Cla & 4.36 ‰ \\
        \midrule
        \multirow{5}[2]{*}{\textbf{5-Hop}} & EV $\to$ RS $\to$ EV $\to$ Aff $\to$ SO & 1.97 ‰ \\
             & EV $\to$ RS $\to$ SO $\to$ Aff $\to$ RS & 1.34 ‰ \\
             & RS $\to$ EV $\to$ SO $\to$ OH $\to$ SO & 0.89 ‰ \\
             & EV $\to$ RS $\to$ ES $\to$ SO $\to$ Aff & 0.45 ‰ \\
             & EV $\to$ ES $\to$ RS $\to$ NE $\to$ Cla & 0.27 ‰ \\
        \bottomrule
    \end{tabular}%
  }
        \vspace{0.5mm}
  \caption{
  Proportions of top-5 strategy transitions in responses. The adjacent same strategies are merged. Abbreviations are consistent with the Appendix B. 
  }
    \vspace{3mm}
  \label{tab:transition}%
\end{table}%

\section{Training Techniques Exploration}
In the rapidly evolving field of artificial intelligence, utilizing large language models in an efficient and effective manner has become increasingly important. Parameter-efficient finetuning stands at the forefront of this pursuit, allowing researchers and practitioners to reuse pre-trained models while minimizing their computational and resource footprints. We explore the following methods to adapt large pre-trained language models on our dataset. We used DialoGPT and LLaMa-7B as the backbones of the compared models:

\subsection{DialoGPT Fine Tuning}
DialoGPT \cite{zhang2020dialogpt} is a medium-sized GPT2 Model trained on 147M conversation-like exchanges extracted from Reddit. It was trained with a causal language modeling (CLM) objective on conversational data and is therefore powerful at response generation in open-domain dialogue systems. In order to fine-tune DialoGPT, we use CLM training. We follow the OpenAI GPT-2\footnote{https://huggingface.co/docs/transformers/model\_doc/gpt2} to model a multiturn dialogue session as a long text and frame the generation task as language modeling.

\begin{table*}[ht]
  \vspace{2mm}
  \centering
  \scalebox{0.85}{
        \begin{tabular}{lcccccccccc}
        \toprule
        \textbf{Backbones} & \textbf{Variants} & \textbf{PPL} & \textbf{METEOR} & \textbf{B-2} & \textbf{B-4} & \textbf{R-L} & \textbf{Extrema} & \textbf{D-1}& \textbf{D-2}& \textbf{D-3}\\
        \midrule
        \multirow{2}[3]{*}{\textbf{DialoGPT}} & no-strategies & 13.11 & 26.03  & 4.438  & 1.721 & 13.37 & 53.27 &19.13 &49.29&62.92\\
    \cmidrule{2-11}          & strategies &  13.71   & 26.82  & 4.773 & 1.966 & 13.23 &55.71  &16.70 &53.11 &77.47 \\
        \midrule
        \multirow{2}[3]{*}{\textbf{LLaMA-Adapter}} & no-strategies & 15.25  & 28.48  & \textbf{6.751} & 1.944 & 16.95 & 64.47 &\textbf{23.23} &60.43 &82.62 \\
    \cmidrule{2-11}          & strategies &  15.82  & 29.71  & 6.317 & 1.987 & 16.39 & 62.73 &22.90 &60.83 &82.24 \\
        \midrule
        \multirow{2}[3]{*}{\textbf{LLaMA-LoRA}} & no-strategies & 15.67 & 30.31 & 6.105 & 2.333 & \textbf{21.60} & 65.06 &21.73 &\textbf{63.64} &\textbf{84.90} \\
    \cmidrule{2-11}          & strategies & \textbf{16.02} & \textbf{30.67}  & 6.416 & \textbf{2.491} & 20.85 & \textbf{65.44} &21.81 &61.94 & 82.80\\
        \bottomrule
        \end{tabular}%
    }
  \caption{
Results of automatic evaluation. The results in bold indicate significant superiority over all the competitors.
  }
    \vspace{2mm}
  \label{tab:automatic}
\end{table*}

\subsection{LLaMA Adapter-Tuning}
LLaMA-Adapter \cite{zhang2023llamaadapter} is a form of prefix-tuning that prepends a learnable adaption-prompt to the inputs of the attention blocks in LLaMA. In total, there are only 1.2M parameters to update during finetuning, which significantly reduces the memory footprint and speeds up training. 
Recently, LLaMA-Adapter v2 \cite{gao2023llamaadapter} is developed to further include more trainable parameters.
We use LLaMA-Adapter v2 to demonstrate instruction-tuning LLaMA 7B on our dataset. Inspired from prefix tuning \cite{li2021prefixtuning} and the original adapter method \cite{houlsby2019}, Adapter-Tuning introduces some new sublayers (i.e., adapter layers) acting as low-rank bottlenecks within each Transformer layer. Generally, instead of tuning all parameters, Adapter-Tuning focuses on tuning mainly the adapter layers. 

\subsection{LLaMA LoRA-Tuning}
Low-rank adaption (LoRA) \cite{hu2021lora} is a technique to approximate the update to the linear layers in a LLM with a low-rank matrix factorization. This significantly reduces the number of trainable parameters and speeds up training with little impact on the final performance of the model. We demonstrate this method by instruction-tuning LLaMA 7B on our dataset. The authors take inspiration from \cite{li2018measuring,aghajanyan-etal-2021-intrinsic} which show that the learned over-parametrized models in fact reside on a low intrinsic dimension. Based on the inherent low-rank characteristics of the large model, the bypass matrix is added to simulate the fine-tuning of the full model parameters. LoRA achieves the purpose of lightweight fine-tuning through a simple and effective solution. It turns various large models into professional models in different fields through light fine-tuning.

\section{Experiments}
In this section, building upon the validation of our ExTES dataset's quality from prior sections, our experiments concentrate on three critical facets: (Q1) Which is the most effective fine-tuning technique for constructing an emotional support chatbot? (Q2) Whether the new dataset adheres to specific criteria, such as toxicity levels, and if it offers broad applicability? (Q3) How does the insights drawn from human assessments complement with automatic evaluations?

\subsection{Automatic Evaluation}

\subsubsection{Experiments on ExTES (Q1)}
To assess the implications of employing support strategies on model performance when using either DialoGPT or LLaMA as the  backbone framework, we evaluated the performance of the no-strategies and strategies versions. The automated evaluation criteria we used comprised of METEOR \cite{banerjeelavie2005meteor}, BLEU-2/4 (B-2/4), ROUGE-L (R-L) \cite{lin2004rouge}, Vector Extrema \cite{Forgues2014} and the Distinct-1/2/3 \cite{li2016diversitypromoting}. The responses were tokenized using the NLTK \cite{loper2002nltk} for this purpose.\par
To demonstrate the effectiveness of the model, we compare it with no-strategies and strategies variants. Results for nine metrics are reported in Table \ref{tab:automatic}. Overall, we have the following observations and discussions:
(1) The models based on LoRA-Tuning and Adapter-Tuning consistently outperform the DialoGPT variant on all metrics, suggesting that LLaMA, with larger parameter size, is more suitable for emotional support chatbot.
(2) The LoRA-Tuning model is slightly better than the Adapter-Tuning model on most metrics (except B-2), indicating that our dataset can play the best effect on LLaMA-LoRA. Therefore, in subsequent experiments, we will use LLaMA-LoRA for evaluation.
(3) The variant with strategies is generally better than the variant without strategies, but slightly lower than the variant without strategies in R-L and D-1/2/3. This is because, under the guidance of strategies, the generation space of replies will be smaller, which reduces the diversity of prediction and generates more targeted replies.

\begin{table}
  \centering
  \renewcommand*{\arraystretch}{1.1}
  \scalebox{0.8}{
    \begin{tabular}{lccc}
    \toprule
    \textbf{Attributes} & \textbf{ExTES} & \textbf{ESConv} & \textbf{LoRA-Responses}\\
    \midrule
    Toxicity & 0.0501 & 0.0760 & \textbf{0.0358} \\
    Severe Toxicity & \textbf{0.0016} & 0.0036 & \textbf{0.0016}\\
    Identify Attack & \textbf{0.0047} & 0.0095 & 0.0048\\
    Insult & 0.0219& 0.0183 & \textbf{0.0137}\\
    Profanity & 0.0251 & 0.0401 & \textbf{0.0222}\\
    Threat & \textbf{0.0073} & 0.0098 & 0.0078\\
    \bottomrule
    \end{tabular}%
  }
  \caption{
    Results of our toxicity assessment using Perspective API. Lower scores are better. The LoRA-Responses are generated by LLaMA-LoRA finetuned on our ExTES dataset.
  }
        \vspace{1mm}
  \label{tab:toxicity}%
\end{table}%

\begin{table*}[!t]
  \centering
  \vspace{3mm}
  \scalebox{0.95}{
    \begin{tabular}{llccccccccc}
    \toprule
    \textbf{Test Set} & \textbf{Train Set} & \textbf{PPL} & \textbf{METEOR} & \textbf{B-2} & \textbf{B-4} & \textbf{R-L} & \textbf{Extrema} & \textbf{D-1}& \textbf{D-2}& \textbf{D-3}\\
    \midrule
    \multirow{2}[3]{*}{ESConv} & ESConv & 15.36  &24.23  & 6.310 & 1.670 & 17.19 & \textbf{58.57} & \textbf{25.72}& 44.09& 60.78\\
\cmidrule{2-11}   & \textbf{ExTES} & \textbf{15.64}  & \textbf{27.07} & \textbf{6.325} & \textbf{2.312} & \textbf{20.57} & 55.56 &24.88 &\textbf{63.83} & \textbf{83.93}\\
    \midrule
    \midrule
    \multirow{2}[3]{*}{ExTES} & ESConv & 15.23  & 24.08 & 6.333 & 1.687 & 16.70 & 53.41  &\textbf{24.79} &46.83 & 65.94\\
\cmidrule{2-11}  & \textbf{ExTES} & \textbf{16.02}   & \textbf{30.67} &  \textbf{6.410}& \textbf{2.491} &\textbf{20.85}& \textbf{65.44}& 21.81&\textbf{61.94} &\textbf{82.80} \\
    \bottomrule
    \end{tabular}%
  }
  \caption{
Experiments across datasets. We apply LoRA-Tuning on LLaMA separately using ESConv and ExTES. The resulting models are then tested on the test set of both datasets for automatic evaluation.
  }
  \vspace{3mm}
  \label{tab:esconv}
\end{table*}

\subsubsection{Toxicity Assessment (Q2)}
It has always been an important yet challenging problem to control language models to avoid generating texts with undesirable attributes, such as toxic language and unnatural repetition \cite{zheng2023click}. We hence use the Perspective API \footnote{\url{https://perspectiveapi.com/}}(a toxicity detection API widely used in online discussions) to evaluate the potential toxicity in our dataset ExTES. All utterances are scored for toxicity across six production attributes via the Perspective API. For each attribute, we report the average score (from 0 to 1, with lower scores being safer) across all utterances.

As shown in Table \ref{tab:toxicity}, our dataset scored very low and exhibited little toxicity. It is generally lower than that of ESConv, the manually curated dataset. We consider the level of toxicity to be normal. Actually, further reductions in toxicity scores may affect the quality of emotional support conversations. Because users seeking emotional support might express some hateful or aggressive contents, which will increase toxicity level in data. It's noteworthy that the score for Severe Toxicity—referring to extremely hateful, aggressive, or disrespectful comments—is as low as 0.0016. This low score can likely be credited to the inherent safety mechanisms within the ChatGPT model. Furthermore, when examining the LLaMA-LoRA model fine-tuned on our ExTES dataset, we observe even reduced levels of toxicity, particularly in categories such as Toxicity, Severe Toxicity, Insult, and Profanity. Such a reduction is advantageous, as our aim is for the emotional support bot to engage users with kindness and courtesy.

\subsubsection{Cross-dataset Experiments (Q2)}
To validate the generality of the collected ExTES dataset, we apply LoRA-Tuning on LLaMA separately using the ESConv and ExTES training set. The resulting models are then tested on the test set of both datasets for comparison.
As shown in Table \ref{tab:esconv}, the performance of the model trained on ESConv is generally worse than the one trained on ExTES across various indicators. 
Firstly, the model trained on ExTES showcases remarkable performance on the ESConv test set, which demonstrates that ExTES possesses remarkable generality to be adapted into various emotional support applications. 
Secondly, the performance gap between the model trained on ExTES and ESConv on the ExTES test set is more substantial than that on the ESConv test set. 
This is mainly because the total amount of ESConv data is small, and there are many unseen scenarios that ESConv does not cover but appears in ExTES test set. 
Naturally, it becomes more difficult to provide an appropriate response for the unseen scenarios. 
Overall, the generality of ExTES can contribute to a more robust and generalizable ES conversation system.

\subsection{Human Evaluation (Q3)}
We performed human interaction evaluation. We recruited 50 students from different majors and collected a total of 100 interactive conversations and ratings (each participant contributed two). Each participant was asked to talk about the same emotional issue with three models. 
Each conversation lasts at least 10 turns, after which participants can continue or end the conversation. Participants are only allowed to talk about emotional issues, but the conversation is open as long as it stays on topic. After the dialogue, participants were asked to rate the performance of the three models according to the following aspects, which followed the evaluation protocol of \cite{huang2021}. To evaluate the models, participants were prompted with the following questions: (1) \textbf{Fluency}: which model's responses were more coherent and easily comprehensible? (2) \textbf{Identification}: which model delved deeper into your situation, effectively identifying your problems? (3) \textbf{Comforting}: which model displayed more adeptness in providing comfort and support? (4) \textbf{Suggestion}: which model offered more helpful suggestions for addressing your issues? (5) \textbf{Overall}: which model's emotional support did you prefer in general? The study encompassed three models: (1) LLaMA-ESConv, (2) LLaMA-ExTES-no strategies and (3) LLaMA-ExTES-strategies.

The outcomes of comparison demonstrate the following findings: (a) It reveals that LoRA-Tuning LLaMa on our ExTES significantly enhanced its capability to provide emotional support across all metrics. 
(b) The strategies version slightly outperforms the non-strategies version overall, especially on Suggestion. This indicates that the strategic guidance is effective and worth further exploration. These findings also underscore the considerable impact of dataset quantity on model performance. 
In general, the fine-tuning of pre-trained models on our dataset rendered them more preferable to users, affirming ExTES's high quality and utility in bolstering emotional support capabilities.

 \begin{table}
    \centering
    \setlength{\tabcolsep}{1.5mm}{
  \scalebox{0.8}{
    \begin{tabular}{lccccccccc}
    \toprule
       \multirow{2}{*}{ExTES vs.}& \multicolumn{3}{c}{ESConv}& \multicolumn{3}{c}{ExTES (no-strategies)}\\
       \cmidrule(lr){2-4}\cmidrule(lr){5-7}
        & Win & Tie & Loss & Win & Tie & Loss  \\
       \midrule
        Flu.  & 28\% & \textbf{50\%} & 22\%   & 43\% & \textbf{48\%} & 9\%\\
        Ide.  & 27\% & \textbf{53\%} & 20\%  & 28\% &\textbf{58\%} &14\% \\
        Com.  & \textbf{49\%} &37\% &14\%    &25\% &28\% &\textbf{47\%} \\
        Sug.  & \textbf{57\%} &22\% &21\%   &\textbf{42\%} &36\% &22\% \\
        Ove.  & \textbf{51\%} &28\% &21\%     & 42\% &\textbf{44\%} &14\% \\
        \bottomrule
    \end{tabular}}
    }
    \caption{Human evaluation results. The results revealed that LoRA-Tuning LLaMa on our ExTES significantly enhanced its capability to provide emotional support across all metrics.}
    \label{tab:human_eval}
\end{table}

\section{Conclusion and Future Work}
In this paper, we address the under-researched integration of emotional support conversation bots with Large Language Models. By overcoming data scarcity and training challenges for buidling emotion support conversation models, we leveraged human expertise and Large Language Models' computational strength to establish a comprehensive and expansive emotional support dialogue dataset. Then we fine-tuned LLaMA to a specialized emotional support chat model. Our method based on extensive emotional support conversation datasets and parameter-efficient tuning, showed promising results in providing emotional support across various scenarios. This study represents a significant advancement in emotional support dialogue systems, laying a strong foundation for future applications and exploration.


\clearpage{
\small
\bibliography{anthology,reference}
\bibliographystyle{acl_natbib}
}

\appendix
\clearpage
\newpage
\appendix

\section{Data Example from our dataset}
\label{app:example}

\begin{figure}[h]
    \centering
    \includegraphics[width=\linewidth]{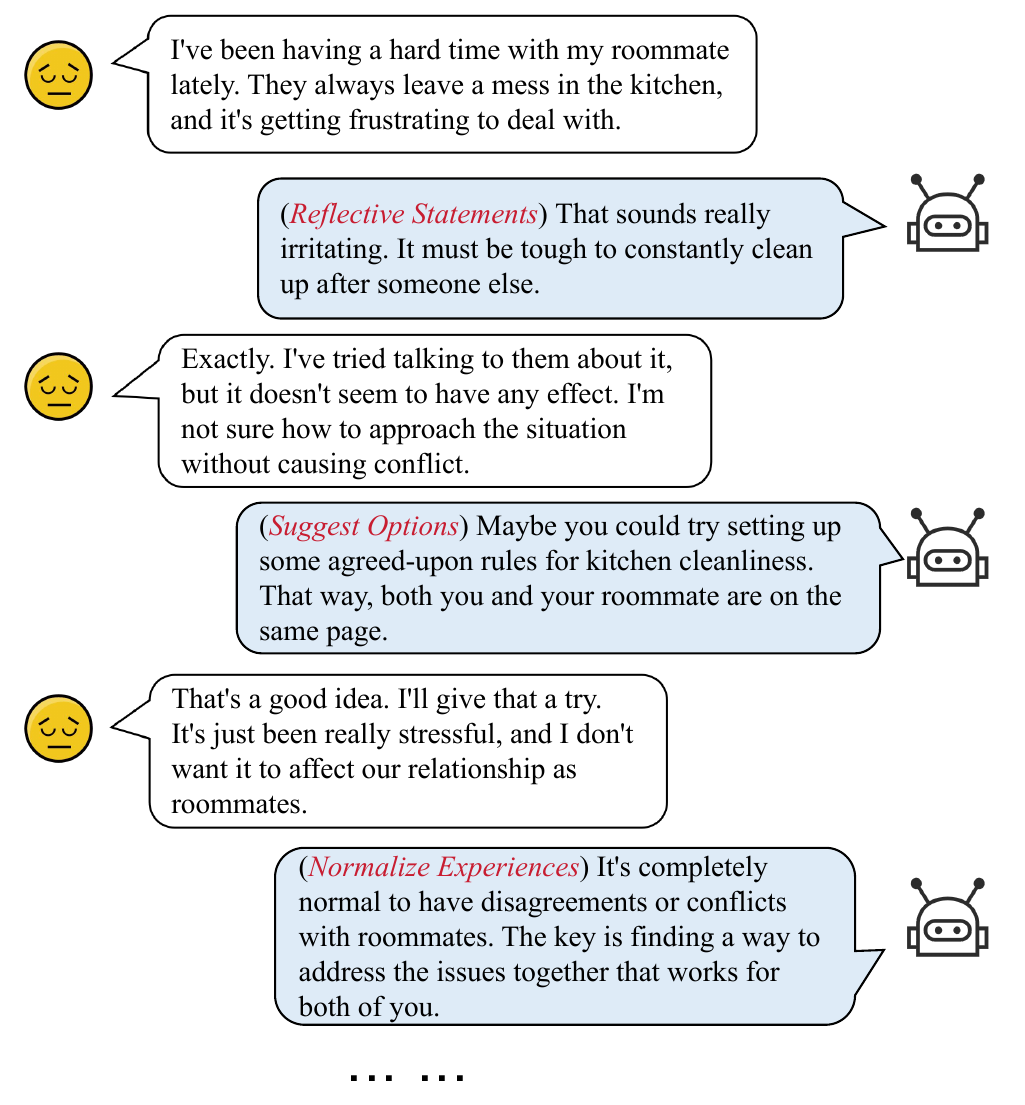}
    \caption{An example chat showing effective emotional support (adapted from our dataset) being provided to the user (left) by the assistant (right). The support strategies used by the assistant are marked in parentheses before the utterances.}
    \label{app_chat_example}
\end{figure}

\begin{figure}[ht]
    \centering
    \includegraphics[width=\linewidth]{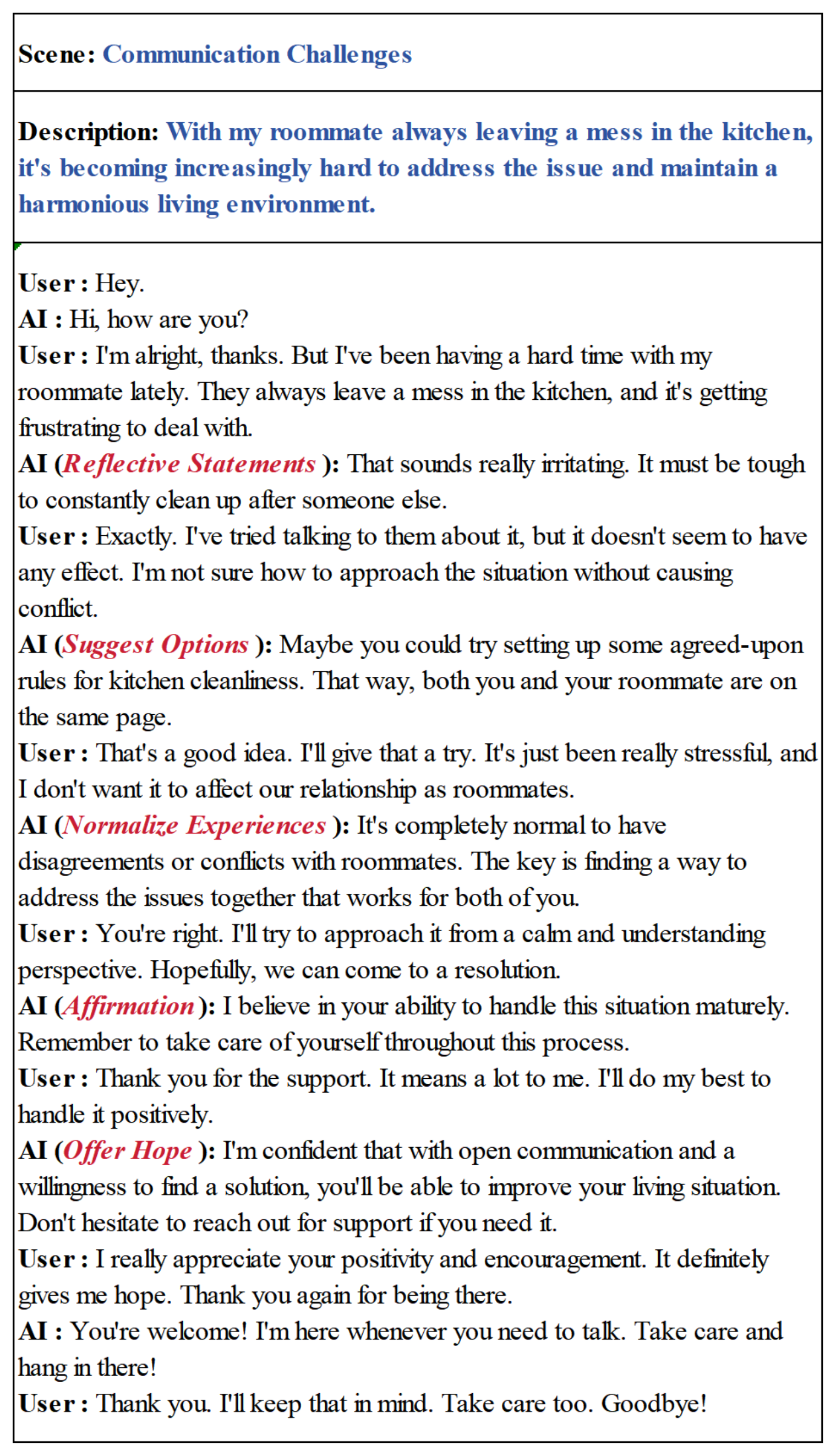}
    \caption{Data example from our dataset.}
    \label{app_dada_example}
\end{figure}

Here we detail the conversation that Figure \ref{app_chat_example} demonstrates to show details that our dataset contains. The detailed example can be seen in Figure \ref{app_dada_example}. Each conversation is labeled its scene category and a brief of description of the user. In the context of each conversation, the strategies used by the assistant are labeled red.

\section{Definitions of Strategies}
\label{sec:definition}
\noindent \textbf{Reflective Statements (RS)}\quad Repeat or rephrase what the User has expressed to show that you're actively listening. Example: User: ``I'm feeling really overwhelmed with all the work I have to do." Assistant: ``It sounds like you're feeling overwhelmed with your workload."

\noindent \textbf{Clarification (Cla)}\quad Seek clarification to ensure a clear understanding of the User's emotions and experiences. Example: User: ``I just can't shake off this feeling of sadness." Assistant: ``Could you help me understand what might have triggered this feeling of sadness?"

\noindent \textbf{Emotional Validation (EV)}\quad Acknowledge and validate the User's emotions without judgment. Example: User: ``I'm so frustrated with myself for making the same mistake again." Assistant: ``It's completely understandable to feel frustrated when you make a mistake."

\noindent \textbf{Empathetic Statements (ES)}\quad Express understanding and empathy towards the User's experiences. Example: User: ``I'm really struggling with my self-confidence right now." Assistant: ``I can imagine how challenging it must be to navigate through situations that affect your self-confidence."

\noindent \textbf{Affirmation (Aff)}\quad Provide positive reinforcement and encouragement to uplift the User's spirits. Example: User: ``I feel like I'm not good enough." Assistant: ``You've accomplished so much already, and your abilities speak for themselves. Don't underestimate your capabilities."

\noindent \textbf{Offer Hope (OH)}\quad Share optimistic perspectives or possibilities to instill hope. Example: User: ``I don't know if things will ever get better." Assistant: ``Remember that change is constant, and there are always opportunities for growth and positive change."

\noindent \textbf{Avoid Judgment and Criticism (AJC)}\quad It's important to create a non-judgmental and safe space for the User to express their emotions without fear of criticism. Refrain from passing judgment or being overly critical of their experiences or choices. Example: User: ``I'm feeling so guilty for taking time off work to focus on my mental health." Assistant: ``Taking care of your mental health is crucial, and it's not something to feel guilty about. Your well-being should always be a priority, and I'm glad you recognized that. Is there anything I can do to support you during this time?"

\noindent \textbf{Suggest Options (SO)}\quad Offer practical suggestions or alternative perspectives for addressing the issue at hand. Example: User: ``I'm having trouble managing my stress." Assistant: ``Have you considered trying relaxation techniques like deep breathing or mindfulness exercises?"

\noindent \textbf{Collaborative Planning (CP)}\quad Work together with the User to develop an action plan. Example: User: ``I want to improve my time management skills." Assistant: ``Let's brainstorm some strategies together. How about breaking tasks into smaller, more manageable chunks?"

\noindent \textbf{Provide Different Perspectives (PDP)}\quad Offer alternative ways of looking at the situation to help the User gain new insights. Example: User: ``I'm devastated that my project didn't succeed." Assistant: ``Sometimes setbacks can lead to unexpected opportunities for learning and growth. It's a chance to reassess and try again."

\noindent \textbf{Reframe Negative Thoughts (RNT)}\quad Help the User reframe negative thoughts into more positive or realistic ones. Example: User: ``I'm such a failure." Assistant: ``Instead of thinking that way, let's focus on what you've learned from this experience and how you can apply it moving forward."

\noindent \textbf{Share Information (SI)}\quad Provide educational or factual information about emotions, coping mechanisms, or self-care practices. Example: User: ``I'm struggling to manage my anxiety." Assistant: ``Did you know that deep breathing exercises and grounding techniques can help reduce anxiety symptoms? Would you like me to explain how to practice them?"

\noindent \textbf{Normalize Experiences (NE)}\quad Explain that certain emotions or reactions are common and part of the human experience. Example: User: ``I feel so guilty for taking time for myself." Assistant: ``It's common to feel guilty about self-care, but it's essential for your well-being. Remember, you deserve to prioritize your needs too."

\noindent \textbf{Promote Self-Care Practices (PSP)}\quad Advocate for engaging in activities that promote well-being and self-care. Example: ``Make sure to take some time for yourself and do something that brings you joy and relaxation."

\noindent \textbf{Stress Management (SM)}\quad Provide suggestions for stress management techniques like exercise, meditation, or spending time in nature. Example: ``Engaging in regular physical activity can help reduce stress and improve mood."

\noindent \textbf{Others (Oth)}\quad Interact with friendly greetings and employ additional supportive techniques that are not covered by the previously mentioned categories.

\section{ChatGPT Self-Chat Template}
\label{sec:template}
The template for ChatGPT self-chat (Figure \ref{Figure1}) is as follows:

\begin{mybox}\tiny\scriptsize\textit{Remember here is a comprehensive list of typical strategies for responding in conversations for emotional support, along with examples for each:
1. Reflective Statements: Repeat or rephrase what the person has expressed to show that you're actively listening.
2. Clarification: Seek clarification to ensure a clear understanding of the person's emotions and experiences.
3. Emotional Validation: Acknowledge and validate the person's emotions without judgment.
4. Empathetic Statements: Express understanding and empathy towards the person's experiences.
5. Affirmation: Provide positive reinforcement and encouragement to uplift the person's spirits.
6. Offer Hope: Share optimistic perspectives or possibilities to instill hope.
7. Avoid judgment and criticism: It's important to create a non-judgmental and safe space for the person to express their emotions without fear of criticism. Refrain from passing judgment or being overly critical of their experiences or choices.
8. Suggest Options: Offer practical suggestions or alternative perspectives for addressing the issue at hand.
9. Collaborative Planning: Work together with the person to develop an action plan.
10. Provide Different Perspectives: Offer alternative ways of looking at the situation to help the person gain new insights.
11. Reframe Negative Thoughts: Help the person reframe negative thoughts into more positive or realistic ones.
12. Share Information: Provide educational or factual information about emotions, coping mechanisms, or self-care practices.
13. Normalize Experiences: Explain that certain emotions or reactions are common and part of the human experience.
14. Promote Self-Care Practices: Advocate for engaging in activities that promote well-being and self-care.
15. Stress Management: Provide suggestions for stress management techniques like exercise, meditation, or spending time in nature.
16. Others: Other strategies.
Example:
\\
\$\{\textbf{SEED EXAMPLE}\}
\\
Your task is to create a casual emotional support conversation between a user and an assistant. Create a random emotional support scenario of the `\$\{\textbf{SCENE}\}' type, write it in the Description, and then generate a complete set of dialogue. Make the conversation more like a real-life chat and be specific. Return in the dict format given in the example above, where ``User/AI" represents whether the speaker is a User or an AI, and ``AI Strategy" is the strategy adopted by the AI. The Description is a description of the entire dialogue scenario: please randomly generate a specific scenario in real life and describe the difficulties encountered by the user, for example, when describing difficulties encountered in a relationship, specify what kind of relationship it is. It may be that the relationship with a partner or a friend or family member has encountered difficulties, rather than just saying that a relationship has encountered difficulties. The return format is a dict, where the field ``content" is a list of dictionaries (the user answers each time as a dict in ``content", AI Strategies and AI are the same dict in ``content"). The ``scene" is the same as in the above example, do not change.}
\end{mybox}

\section{Examples of Scenarios}
\label{sec:scene}

\noindent \textbf{Breakups or Divorce}\quad Example 1: Processing the emotions and grief following the end of a long-term relationship. Example 2: Seeking guidance on how to navigate a recent breakup and move forward.

\noindent \textbf{Conflicts or Communication Problems}\quad Example 1: Dealing with a misunderstanding or disagreement with a close friend or family member. Example 2: Seeking advice on resolving conflicts with a romantic partner and improving communication.

\noindent \textbf{Communication Challenges}\quad Example: Helping a person find effective ways to express their needs and concerns to their partner, fostering open and constructive communication.

\noindent \textbf{Coping with the Death of a Loved One}\quad Example 1: Navigating the stages of grief and finding ways to honor the memory of the deceased. Seeking support in managing the emotional impact of losing a close family member or friend.

\noindent \textbf{Dealing with the Loss of a Pet}\quad Example 1: Processing the deep sadness and emptiness after the death of a beloved pet. Example 2: Seeking understanding and comfort while grieving the loss of a long-time companion animal.

\noindent \textbf{Work-related Stress and Burnout}\quad Example 1: Coping with excessive workload, pressure, and a demanding work environment. Example 2: Seeking strategies to manage stress and achieve a healthier work-life balance.

\noindent \textbf{Financial Worries and Uncertainty}\quad Example 1: Navigating financial challenges such as debt, job loss, or unexpected expenses. Example 2: Seeking emotional support and practical advice to alleviate financial stress and regain stability.
\noindent \textbf{Unemployment-related stress}\quad Example: Encouraging someone who is about to lose their job due to poor company performance, discussing the possibility of changing jobs, prioritizing self-care, and staying positive.

\noindent \textbf{Academic Stress}\quad Example: Offering guidance and study tips to a student feeling overwhelmed by their workload, helping them create a study plan and adopt healthy stress management techniques.

\noindent \textbf{Depression and Low Mood}\quad Example 1: Dealing with feelings of sadness, loss of interest, and lack of motivation. Example 2: Seeking guidance on coping mechanisms and professional help for managing depression symptoms.

\noindent \textbf{Managing Bipolar Disorder}\quad Example 1: Finding support and strategies to navigate the highs and lows of bipolar disorder. Example 2: Seeking advice on maintaining stability, managing medication, and recognizing warning signs.

\noindent \textbf{Anxiety and Panic}\quad Example: Providing guidance and techniques for someone who experiences social anxiety, helping them gradually face their fears and build confidence in social situations.

\noindent \textbf{Depression and Low Mood}\quad Example: Being there for a person experiencing depression, actively listening to their struggles, and encouraging them to seek professional help and engage in self-care activities.

\noindent \textbf{Adjusting to a New Job or Role}\quad Example 1: Coping with the challenges and expectations of a new job or promotion. Example 2: Seeking guidance on adapting to a new work environment and building professional relationships.

\noindent \textbf{Moving to a New City or Country}\quad Example 1: Dealing with feelings of homesickness, cultural adjustment, and building a new social network. Example 2: Seeking support in navigating the practical and emotional aspects of relocating to a different city or country.

\noindent \textbf{Career Transitions}\quad Example: Assisting someone who is considering a career change, helping them explore their passions, transferable skills, and develop a plan for transitioning into a new field.

\noindent \textbf{Parenthood and Parenting Challenges}\quad Example: Supporting a new parent who is feeling overwhelmed and sleep-deprived, offering reassurance, and sharing tips for self-care and coping strategies for the demands of parenthood.

\noindent \textbf{Low Self-Esteem or Lack of Confidence}\quad Example 1: Addressing negative self-perceptions and building self-worth. Example 2: Seeking techniques for cultivating self-compassion and improving self-esteem.

\noindent \textbf{Body Image Concerns and Eating Disorders}\quad Example 1: Dealing with body dissatisfaction and the impact it has on self-image and overall well-being. Example 2: Seeking support in recovering from an eating disorder and developing a healthy relationship with food and body.

\noindent \textbf{LGBTQ+ Identity}\quad Example: Assisting someone in the process of coming out as gay, offering support, connecting them with LGBTQ+ community resources, and being a source of understanding.

\noindent \textbf{Cultural Identity and Belonging}\quad Example: Engaging in discussions with someone who is exploring their mixed-race identity and helping them embrace and celebrate their diverse heritage.

\noindent \textbf{Academic Stress or Pressure}\quad Example 1: Coping with academic expectations, exam anxiety, or perfectionism. Example 2: Seeking strategies for time management, study techniques, and reducing academic stress.

\noindent \textbf{Job Loss or Career Setbacks}\quad Example 1: Navigating the emotions and challenges of losing a job or facing career setbacks. Example 2: Seeking guidance and encouragement for career transitions or exploring new professional opportunities.

\noindent \textbf{Parenting Challenges and Parental Guilt}\quad Example 1: Managing parental responsibilities, parenting styles, and dealing with parental guilt. Example 2: Seeking advice on effective communication with children and finding a balance between work and family.

\noindent \textbf{Sibling Rivalry or Family Conflict}\quad Example 1: Resolving conflicts and improving relationships with siblings or other family members. Example 2: Seeking guidance on navigating family dynamics, establishing healthy boundaries, and fostering understanding.

\noindent \textbf{Surviving and Recovering from Physical or Emotional Abuse}\quad Example 1: Processing the trauma of past abuse and seeking support for healing and recovery. Example 2: Finding resources and coping strategies for managing the emotional impact of abuse.

\noindent \textbf{Healing from Sexual Assault or Domestic Violence}\quad Example 1: Navigating the complex emotions, seeking support, and developing coping mechanisms after experiencing sexual assault or domestic violence. Example 2: Accessing information on trauma-informed therapy and support networks for survivors of assault or violence.

\noindent \textbf{Post-Traumatic Stress Disorder (PTSD)}\quad Example: Creating a safe and non-judgmental space for a military veteran with PTSD to share their experiences and providing resources for trauma-focused therapy and support groups.

\noindent \textbf{Healing from Abuse}\quad Example: Assisting someone who has recently left an abusive relationship, connecting them with local support services, and offering encouragement as they rebuild their life.

\noindent \textbf{Navigating Gender Identity and Transitioning}\quad Example 1: Seeking support and resources while exploring gender identity and considering transitioning. Example 2: Accessing guidance on navigating social, medical, and legal aspects of transitioning.

\noindent \textbf{Chronic Illness or Pain Management}\quad Example 1: Coping with the emotional impact of a chronic illness, including pain, limitations, and lifestyle adjustments. Example 2: Seeking support in managing daily challenges, finding self-care strategies, and connecting with others facing similar health issues.

\noindent \textbf{Coping with a Diagnosis or Medical Treatment}\quad Example 1: Processing the emotions surrounding a new medical diagnosis and navigating treatment options. Example 2: Seeking emotional support and practical guidance to cope with medical procedures, side effects, and lifestyle changes.

\noindent \textbf{Caregiver Support}\quad Example: Offering guidance and resources to a caregiver of an elderly parent, discussing techniques for managing caregiver stress and suggesting respite care options.

\noindent \textbf{Finding Meaning and Purpose in Life}\quad Example 1: Exploring questions related to the meaning of life, personal values, and finding purpose. Example 2: Assisting someone who is questioning their life's purpose and exploring different avenues for finding meaning, discussing their values and interests, and encouraging self-reflection.

\noindent \textbf{Spirituality and Faith}\quad Example: Offering guidance and resources to someone who is questioning their faith or seeking spiritual fulfillment, providing support ass they explore their beliefs and values.

\noindent \textbf{Addiction and Recovery}\quad Example: Offering empathy and understanding to someone battling addiction, discussing treatment options, and providing emotional support during their journey to recovery.

\noindent \textbf{Support for Loved Ones or Friends}\quad Example: Supporting a parent who has a child dealing with addiction, offering a listening ear, and connecting them with support groups and counseling services.

\end{document}